\documentclass[10pt,twocolumn,letterpaper]{article}

\usepackage{cvpr}
\usepackage{times}
\usepackage{epsfig}
\usepackage{graphicx}
\usepackage{amsmath}
\usepackage{amssymb}
\usepackage{booktabs}

\usepackage[pagebackref=true,breaklinks=true,letterpaper=true,colorlinks,bookmarks=false]{hyperref}

\cvprfinalcopy 

\def\cvprPaperID{2525} 

\ifcvprfinal\fi
\begin{document}

\title{Non-Adversarial Image Synthesis with Generative Latent Nearest Neighbors}

\author{Yedid Hoshen\\ 
Facebook AI Research\\
\and
Jitendra Malik\\
Facebook AI Research and UC Berkeley
}

\maketitle

\begin{abstract}
   Unconditional image generation has recently been dominated by generative adversarial networks (GANs). GAN methods train a generator which regresses images from random noise vectors, as well as a discriminator that attempts to differentiate between the generated images and a training set of real images. GANs have shown amazing results at generating realistic looking images. Despite their success, GANs suffer from critical drawbacks including: unstable training and mode-dropping. The weaknesses in GANs have motivated research into alternatives including: variational auto-encoders (VAEs), latent embedding learning methods (e.g. GLO) and nearest-neighbor based implicit maximum likelihood estimation (IMLE). Unfortunately at the moment, GANs still significantly outperform the alternative methods for image generation. In this work, we present a novel method - Generative Latent Nearest Neighbors (GLANN) - for training generative models without adversarial training. GLANN combines the strengths of IMLE and GLO in a way that overcomes the main drawbacks of each method. Consequently, GLANN generates images that are far better than GLO and IMLE. Our method does not suffer from mode collapse which plagues GAN training and is much more stable. Qualitative results show that GLANN outperforms a baseline consisting of 800 GANs and VAEs on commonly used datasets. Our models are also shown to be effective for training truly non-adversarial unsupervised image translation.          
\end{abstract}

\section{Introduction}
\label{sec:intro}

Generative image modeling is a long-standing goal for computer vision. Unconditional generative models set to learn functions that generate the entire image distribution given a finite number of training samples. Generative Adversarial Networks (GANs) \cite{goodfellow2014generative} are a recently introduced technique for image generative modeling. They are used extensively for image generation owing to: i) training effective unconditional image generators ii) being almost the only method for unsupervised image translation between domains (but see NAM \cite{nam_eccv}) iii) being an effective perceptual image loss function (e.g. Pix2Pix \cite{pix2pix}).

Along with their obvious advantages, GANs have critical disadvantages: i) GANs are very hard to train, this is expressed by a very erratic progression of training, sudden run collapses, and extreme sensitivity to hyper-parameters. ii) GANs suffer from mode-dropping - the modeling of only some but not all the modes of the target distribution. The birthday paradox can be used to measure the extent of mode dropping \cite{birthday}: The number of modes modeled by a generator can be estimated by generating a fixed number of images and counting the number of repeated images. Empirical evaluation of GANs found that the number of modes is significantly lower than the number in the training distribution.

The disadvantages of GANs gave rise to research into non-adversarial alternatives for training generative models. GLO \cite{glo} and IMLE \cite{imle} are two such methods. GLO, introduced by Bojanowski et al., embeds the training images in a low dimensional space, so that they are reconstructed when the embedding is passed through a jointly trained deep generator. The advantages of GLO are i) encoding the entire distribution without mode dropping ii) the learned latent space corresponds to semantic image properties i.e. Euclidean distances between latent codes correspond to semantically meaningful differences. A critical disadvantage of GLO is that there is not a principled way to sample new images from it. Although the authors’ recommended fitting a Gaussian to the latent codes of the training images, this does not result in high-quality image synthesis.

\begin{figure*}
  \centering
\includegraphics[width=0.85\linewidth]{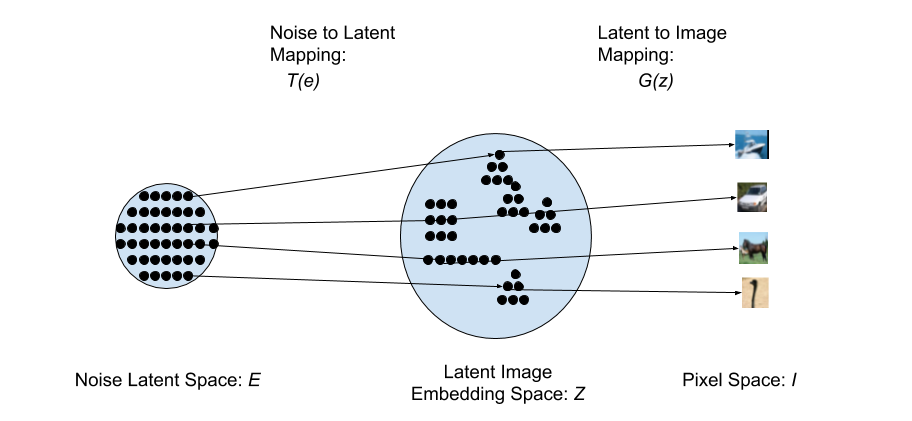}

  \caption{An illustration of our architecture: a random noise vector $e$ is sampled and mapped to the latent space to yield latent code $z=T(e)$. The latent code is projected by the generator to yield image $I = G(z)$.}
  \label{fig:gen_shoes}
\end{figure*}

IMLE was proposed by Li and Malik \cite{imle} for training generative models by sampling a large number of latent codes from an arbitrary distribution, mapping each to the image domain using a trained generator and ensuring that for every training image there exists a generated image which is near to it. IMLE is trivial to sample from and does not suffer from mode-dropping. Like other nearest neighbor methods, IMLE is sensitive to the exact metric used, particularly given that the training set is finite.  Recall that while the classic Cover-Hart result \cite{cover1967nearest} tells us that asymptotically the error rate of the nearest neighbor classifier is within a factor of 2 of the Bayes risk, when we use a finite set of exemplars better choices of metrics give us better classifier performance. When trained directly on image pixels using an $L_2$ loss, IMLE synthesizes blurry images.

In this work, we present a new technique, Generative Latent Nearest Neighbors (GLANN), which is able to train generative models of comparable or better quality to GANs. Our method overcomes the metric problem of IMLE by first embedding the training images using GLO. The attractive linear properties of the latent space induced by GLO, allow the Euclidean metric to be semantically meaningful in the latent space $\cal Z$. We train an IMLE-based model to map between an arbitrary noise distribution $\cal E$, and the GLO latent space $\cal Z$. The GLO generator can then map the generated latent codes to pixel space, thus generating an image. Our method GLANN enjoys the best of both IMLE and GLO: easy sampling, modeling the entire distribution, stable training and sharp image synthesis. A schema of our approach is presented in Fig.~\ref{fig:gen_shoes}.

We quantitatively evaluate our method using established protocols and find that it significantly outperforms other non-adversarial methods, while being usually better or competitive with current GAN based models. GLANN is also able to achieve promising results on high-resolution image generation and 3D generation. Finally, we show that GLANN-trained models are the first to perform truly non-adversarial unsupervised image translation.

\section{Previous Work}
\label{sec:prev}

\textbf{Generative Modeling:} Generative modeling of images is a long-standing problem of wide applicability. Early approaches included mixtures of Gaussian models (GMM) \cite{zoran2011learning}. Such methods were very limited in image resolution and quality. Since the introduction of deep learning, deep methods have continually been used for image generative models. Early attempts included Deep Belief Networks (DBNs) (e.g. \cite{bengio2013better}). DBNs however were rather tricky to train and did not scale to high resolutions. Variational Autoencoders (VAEs) \cite{vae} are a significant breakthrough in deep generative modeling, introduced by Kingma and Welling. VAEs are able to generate images from the Gaussian distribution by making a variational approximation. This scheme was followed by multiple works including the recent Wasserstein Autoencoder \cite{tolstikhin2017wasserstein}. Although VAEs are relatively simple to train and have solid theoretical foundations, they generally do not generate sharp images. This is partially due to making restrictive assumptions such as a unimodal prior and requirement for an encoder. 

Several other non-adversarial training paradigms exist: Generative invertible flows \cite{dinh2014nice}, that were recently extended to high resolution \cite{kingma2018glow} but at prohibitive computational costs. Another training paradigm is autoregressive image models e.g. PixelRNN/PixelCNN \cite{oord2016pixel}, where pixels are modeled sequentially. Autoregressive models are computationally expensive and underperform adversarial methods although they are the state of the art in audio generation (e.g. WaveNet \cite{oord2016wavenet}).   

\textbf{Adversarial Generative Models:} Generative Adversarial Networks (GANs) were first introduced by Goodfellow et al. \cite{goodfellow2014generative} and are the state-of-the-art method for training generative models. A basic discussion on GANs was given in Sec.~\ref{sec:intro}. GANs have shown a remarkable ability for image generation, but suffer from difficult training and mode dropping. Many methods were proposed for improving GANs e.g. changing the loss function (e.g. Wasserstein GAN \cite{arjovsky2017wasserstein}) or regularizing the discriminator to be Lipschitz by: clipping \cite{arjovsky2017wasserstein}, gradient regularization \cite{gulrajani2017improved, Mescheder2018ICML} or spectral normalization \cite{miyato2018spectral}. GAN training was shown to scale to high resolutions \cite{zhang2018self} using engineering tricks and careful hyper-parameter selection.

\textbf{Evaluation of Generative Models:} Evaluation of generative models is challenging. Early works evaluated generative models using probabilistic criteria (e.g. \cite{zoran2011learning}). More recent generative models (particularly GANs) are not amenable to such evaluation. GAN generations have traditionally been evaluated using visual inspection of a handful of examples or by a user study. More recently, more principled evaluation protocols have emerged. Inception Scores (IS) which take into account both diversity and quality were first introduced by \cite{salimans2016improved}. FID scores \cite{heusel2017gans} were more recently introduced to overcome major flaws of the IS protocol \cite{barratt2018note}. Very recently, a method for generative evaluation which is able to capture both precision and recall was introduced by Sajjadi et al. \cite{sajjadi2018assessing}. Due to the hyperparameters sensitivity of GANs, a large scale study of the performance of $7$ different GANs and VAE was carried out by Lucic et al. \cite{lucic2017gans} over a large search space of 100 different hyperparameters, establishing a common baseline for evaluation.

\textbf{Non-Adversarial Methods:} The disadvantages of GANs motivated research into GAN alternatives. GLO \cite{glo}, a recently introduced encoder-less generative model which uses a non-adversarial loss function, achieves better results than VAEs. Due to the lack of a good sampling procedure, it does not outperform GANs (see Sec.~\ref{sec:method:glo}). IMLE \cite{imle}, a method related to ICP was also introduced for training unconditional generative models, however due to computational challenges and the choice of metric, it also does not outperform GANs. Chen and Koltun \cite{chen2017photographic} presented a non-adversarial method for supervised image mapping, which in some cases was found to be competitive with adversarial methods. Hoshen and Wolf introduced an ICP-based method \cite{mbcicp} for unsupervised word translation which contains no adversarial training. However, this method is not currently able to generate high quality images. They also presented non-adversarial method, NAM \cite{nam_iclr,nam_eccv,vaenam}, for unsupervised image mapping. The method relies on having access to a strong unconditional model of the target domain, which is typically trained using GANs.      

\section{Our method}
\label{sec:method}

In this section we present a method - GLANN - for synthesizing high-quality images without using GANs.

\subsection{GLO}
\label{sec:method:glo}

Classical methods often factorize a set of data points $\{x_1,x_2,..,x_T\}$ via the following decomposition:
\begin{equation}
x_i = Wz_i~~~~~\forall{i}
\end{equation}

Where $z_i$ is a latent code describing $x_i$, and $W$ is a set of weights. Such factorization is poorly constrained and is typically accompanied by other constraints such as low-rank, positivity (NMF), sparsity etc. Both $W$ and $z_i$ are optimized directly e.g. by alternating least squares or SVD. The resulting $z_i$ are latent vectors that embed the data in a lower dimension and typically better behaved space. It is often found that attributes become linear operations in the latent space. 

GLO \cite{glo} is a recently introduced deep method, which is different from the above in three aspects: i) Constraining all latent vectors to lie on a unit sphere or a unit ball. ii) Replacing the linear matrix $W$, by a deep CNN generator $G()$ which is more suitable for modeling images. iii) Using a Laplacian pyramid loss function (but we find that a VGG \cite{simonyan2014very} perceptual loss works better). 

The GLO optimization objective is written in Eq.~\ref{eq:glo_def}:

\begin{equation}
\label{eq:glo_def}
arg\min_{G, \{z_i\}}\sum_i\ell(G(z_i), x_i) ~~~s.t.~~~~ \|z_i\|=1
\end{equation}

Bojanowski et al \cite{glo}, implement $\ell$ as a Laplacian pyramid. All weights  are trained by SGD (including the generator weights $G()$ and a latent vector $z_i$ per each training image $x_i$). After training, the result is a generator $G()$ and a latent embedding $z_i$ of each training image $x_i$. 

\subsection{IMLE}
\label{sec:method:imle}
IMLE \cite{imle} is a recent non-adversarial technique that maps between distributions using a maximum likelihood criterion. Each epoch of IMLE consists of the following stages: i) $M$ random latent codes $e_j$ are sampled from a normal distribution ii) The latent codes are mapped by the generator resulting in images $G(e_j)$ iii) For each training example $x_i$, the nearest generated image is found such that: $e_i = arg\min_{e_j}\|G(e_j), x_i|^2_2$ iv) $G()$ is optimized using nearest neighbors as approximate correspondences  $G = arg\min_{\tilde{G}} \sum_{i}{\|\tilde{G}(e_i), x_i\|^2_2}$
This procedure is repeated until the convergence of $G()$.

\subsection{Limitations of GLO and IMLE}
\label{sec:method:limit}
The main limitation of GLO is that the generator is not trained to sample from any known distribution i.e. the distribution of $z_i$ is unknown and we cannot directly sample from it. When sampling latent variables from a normal distribution or when fitting a Gaussian to the training set latent codes (as advocated in \cite{glo}), generations that are of much lower quality than GANs are usually obtained. This prevents GLO from being competitive with GANs.

Although sampling from an IMLE trained generator is trivial, the training is not, a good metric might not be known, the nearest neighbor computation and feature extraction for each random noise generation is costly. IMLE typically results in blurry image synthesis.

\subsection{GLANN: Generative Latent Nearest Neighbor}
\label{sec:method:glann}

We present a method - GLANN - that overcomes the weaknesses of both GLO and IMLE. GLANN consists of two stages: i) embedding the high-dimensional image space into a "well-behaved" latent space using GLO. ii) Mapping between an arbitrary distribution (typically a multi-dimensional normal distribution) and the low-dimensional latent space using IMLE.

\subsubsection{Stage 1: Latent embedding}
\label{sec:method:glann:glo}

Images are high-dimensional and distances between them in pixel space might not be meaningful. This makes IMLE and the use of simple metric functions such as $L_1$ or $L_2$ less effective in pixel space.  In some cases perceptual features may be found under which distances make sense, however they are  high dimensional and expensive to compute.

Instead our method first embeds the training images in a low dimensional space using GLO. Differently from the GLO algorithm, we use a VGG perceptual loss function. The optimization objective is written in Eq,~\ref{eq:ours:lowdim}:  

\begin{equation}
\label{eq:ours:lowdim} 
arg\min_{\tilde{G}, \{z_i\}}\sum_i \ell_{perceptual}(\tilde{G}(z_i), x_i) ~~~~ s.t. ~~~~ \|z_i\|=1
\end{equation}

All parameters are optimized directly by SGD. By the end of training, the training images are embedded by the low dimensional latent codes $\{z_i\}$. The latent space $\cal Z$ enjoys convenient properties such as linearity. A significant benefit of this space is that a Euclidean metric in the $\cal Z$ space can typically yield more more semantically meaningful results than raw image pixels.

\subsubsection{Stage 2: Sampling from the latent space}
\label{sec:method:glann:imle}

GLO replaced the problem of sampling from image pixels $\cal X$ by the problem of sampling from $\cal Z$ without offering an effective sampling algorithm. Although the original paper suggests fitting a Gaussian to the training latent vectors $z_i$, this typically does not result in good generations. Instead we propose learning a mapping from a distribution from which sampling is trivial (e.g. multivariate normal) to the empirical latent code distribution using IMLE.

At the beginning of each epoch, we sample a set of random noise codes $e_1..e_m..e_M$ from the noise distribution. Each one of the codes is mapped using mapping function $T$ to the latent space:

\begin{equation}
\label{eq:ours:lowdim} 
\tilde{z}_m = T(e_m)
\end{equation}

During the epoch, our method iteratively samples a minibatch of latent codes  from the set $\{z_1..z_t..z_T\}$ computed in the previous stage. For each latent code $z_t$, we find the nearest neighbor mapped noise vector (using a Euclidean distance metric):

\begin{equation}
\label{eq:ours:lowdim} 
e_t = \arg\min_{e_m} \|z_t - T(e_m)\|^2_2
\end{equation}

The approximate matches can now be used for finetuning the mapping function $T$:

\begin{equation}
\label{eq:ours:lowdim} 
T = \arg\min_{\tilde{T}}\sum_t \|z_t - \tilde{T}(e_t)\|^2_2
\end{equation}

This procedure is repeated until the convergence of $T()$. It was shown theoretically by Li and Malik \cite{imle}, that the method achieves a form of maximum likelihood estimate.

\subsubsection{Sampling new images}
\label{sec:method:lamlis:sampling}

Synthesizing new images is now a simple task: We first sample a noise vector from the multivariate normal distribution $e \sim N(0, I)$. The new sample is mapped to the latent code space:
\begin{equation}
\label{eq:ours:sample_latent} 
z_e = T(e)
\end{equation}

By our previous optimization, $T()$ was trained such that latent code $z_e$ lies close to the data manifold. We can therefore use the generator to project the latent code to image space by our GLO trained generator $G()$:

\begin{equation}
\label{eq:ours:sample_pixel} 
I_e = G(z_e)
\end{equation}

$I_e$ will appear to come from the distribution of the input images $x$.

It is also possible to invert this transformation by optimizing for the noise vector $e$ given an image $I$:

\begin{equation}
\label{eq:ours:sample_invert} 
e = arg\min_{\tilde{e}}\ell(G(T(\tilde{e})), I)
\end{equation}

\begin{table*}[t]
  \centering
      
  \caption{Quality of Generation (FID)}
  \label{tab:lucic}

    \begin{tabular}{lcccccccc}
    \toprule
    & && Adversarial & & & & Non-Adversarial & \\
    \cmidrule(l){1-1} \cmidrule(l){2-6} \cmidrule(l){7-9}
    Dataset & MM GAN & NS GAN & LSGAN & WGAN & BEGAN & VAE & GLO & Ours \\ 
    \cmidrule(l){1-1} \cmidrule(l){2-9}
   MNIST  & $9.8 \pm 0.9$ & $6.8 \pm 0.5$ & $7.8 \pm 0.6$& \textbf{6.7} $\pm$  0.4 &  $13.1 \pm 1.0$ & $23.8 \pm 0.6$ & $49.6 \pm 0.3$ & $8.6 \pm 0.1$ \\
   Fashion  & $29.6 \pm 1.6$ & $26.5 \pm 1.6$& $30.7 \pm 2.2$& $21.5 \pm 1.6$& $22.9 \pm 0.9$ & $58.7 \pm 1.2$ & $57.7 \pm 0.4$& $\textbf{13.0} \pm 0.1$ \\
   Cifar10  & $72.7 \pm 3.6$ & $58.5 \pm 1.9$ & $87.1 \pm 47.5$& $55.2 \pm 2.3$& $71.4 \pm 1.6$& $155.7 \pm 11.6$ & $65.4 \pm 0.2$ & $\textbf{46.5} \pm 0.2$ \\
   CelebA  & $65.6 \pm 4.2$ & $55.0 \pm 3.3$ & $53.9 \pm 2.8$& $41.3 \pm 2.0$& $\textbf{38.9} \pm 0.9$& $85.7 \pm 3.8$ & $52.4 \pm 0.5$ &  $46.3 \pm 0.1$ \\
	 \bottomrule
    \end{tabular}
\end{table*}

\section{Experiments}

To evaluate the performance of our proposed method, we perform quantitative and qualitative experiments comparing our method against established baselines.

\subsection{Quantitative Image Generation Results}
\label{sec:eval:fid}

In order to compare the quality of our results against representative adversarial methods, we evaluate our method using the protocol established by Lucic et al.~\cite{lucic2017gans}. This protocol fixes the architecture of all generative models to be InfoGAN \cite{infogan}. They evaluate $7$ representative adversarial models (DCGAN, LSGAN, NSGAN, W-GAN, W-GAN GP, DRAGAN, BEGAN) and a single non-adversarial model (VAE). In \cite{lucic2017gans}, significant computational resources are used to evaluate the performance of each method over a set of 100 hyper-parameter settings, e.g.: learning rate, regularization, presence of batch norm etc. 

Finding good evaluation metrics for generative models is an active research area. Lucic et al. argue that the previously used Inception Score (IS) is not a good evaluation metric, as the maximal IS score is obtained by synthesizing a single image from every class. Instead, they advocate using Frechet Inception Distance (FID) \cite{heusel2017gans}. FID measures the similarity of the distributions of real and generated images by two steps: i) Running the Inception network as a feature extractor to embed each of the real and generated images ii) Fitting a multi-variate Gaussian to the real and generated embeddings separately, to yield means $\mu_r$, $\mu_g$ and variances $\Sigma_r$, $\Sigma_g$ for the real and generated distributions respectively. The FID score is then computed as in Eq.~\ref{eq:method:fid}:

\begin{equation}
\label{eq:method:fid} 
FID = \|\mu_r - \mu_g\|^2_2 + Tr(\Sigma_r + \Sigma_g - 2(\Sigma_r \Sigma_g)^{\frac{1}{2}})
\end{equation}

Lucic et al. evaluate the $8$ baselines on $4$ standard public datasets: MNIST \cite{mnist}, Fashion MNIST \cite{xiao2017fashion}, CIFAR10 \cite{cifar} and CelebA \cite{celeba}. MNIST, Fashion-MNIST and CIFAR10 contain 50k color images and 10k validation images. MNIST and Fashion are $28 \times 28$ while CIFAR is $32 \times 32$.

For a fair comparison of our method, we use the same generator architecture used by Lucic et al. for our GLO model. We do not have a discriminator, instead, we use a VGG perceptual loss. Also differently from the methods tested by Lucic et al. we train an additional network $T()$ for IMLE sampling from the noise space to the latent space. In our implementation, $T()$ has two dense layers with $128$ hidden nodes, with RelU and BatchNorm. GLANN actually uses fewer parameters than the baseline by not using a discriminator. Our method was trained with ADAM \cite{adam}. We used the highest learning rate that allowed convergence: $0.001$ for the mapping network, $0.01$ for the latent codes ($0.003$ for CelebA), generator learning rate was $0.1 \times$ the latent code rate. $500$ epochs were used for GLO training decayed by $0.5$ every 50 epochs. $50$ epochs were used for mapping network training.


Tab.~\ref{tab:lucic} presents a comparison of the FID achieved by our method and those reported by Lucic et al. We removed DRAGAN and WGAN-GP for space consideration (and as other methods represented similar performance). The results for GLO were obtained by fitting a Gaussian to the learned latent codes (as suggested in \cite{glo}).

On Fashion and CIFAR10, our method significantly outperforms all baselines - despite just using a single hyper-parameter setting. Our method is competitive on MNIST, although it does not reach the top performance. As most methods performed very well on this task, we do not think that it has much discriminative power. We found that a few other methods outperformed ours in terms of FID on CelebA, due to checkerboard patterns in our generated images. This is a well known phenomenon of deconvolutional architectures \cite{odena2016deconvolution}, which are now considered outdated. In Sec.~\ref{sec:eval:qual}, we show high-quality CelebA-HQ facial images generated by our method when trained using modern architectures. 

Our method always significantly outperforms the VAE and GLO baseline, which are strong representatives of non-adversarial methods. One of the main messages in \cite{lucic2017gans} was that GAN methods require a significant hyperparameter search to achieve good performance. Our method was shown to be very stable and achieved strong performance (top on two datasets) with a fixed hyperparameter setting. An extensive hyperparameter search can potentially further increase the performance our method, we leave it to future work.

\begin{figure*}
  \centering
    \begin{tabular}{cccc}
MNIST & Fashion & CIFAR10 & CelebA \\
\includegraphics[width=0.24\linewidth]{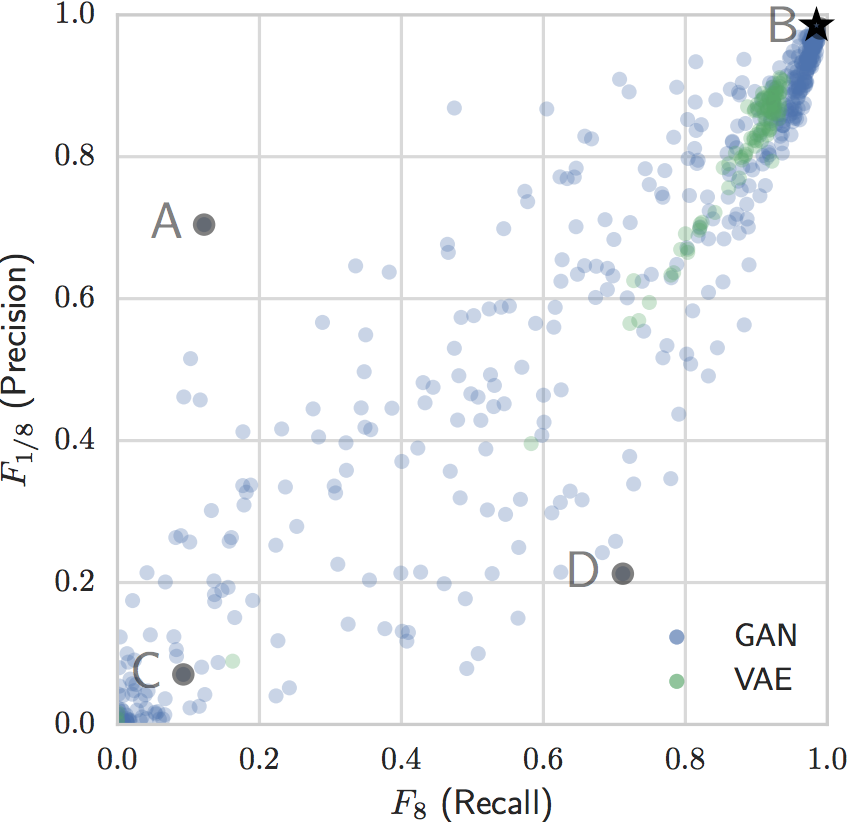} &
\includegraphics[width=0.24\linewidth]{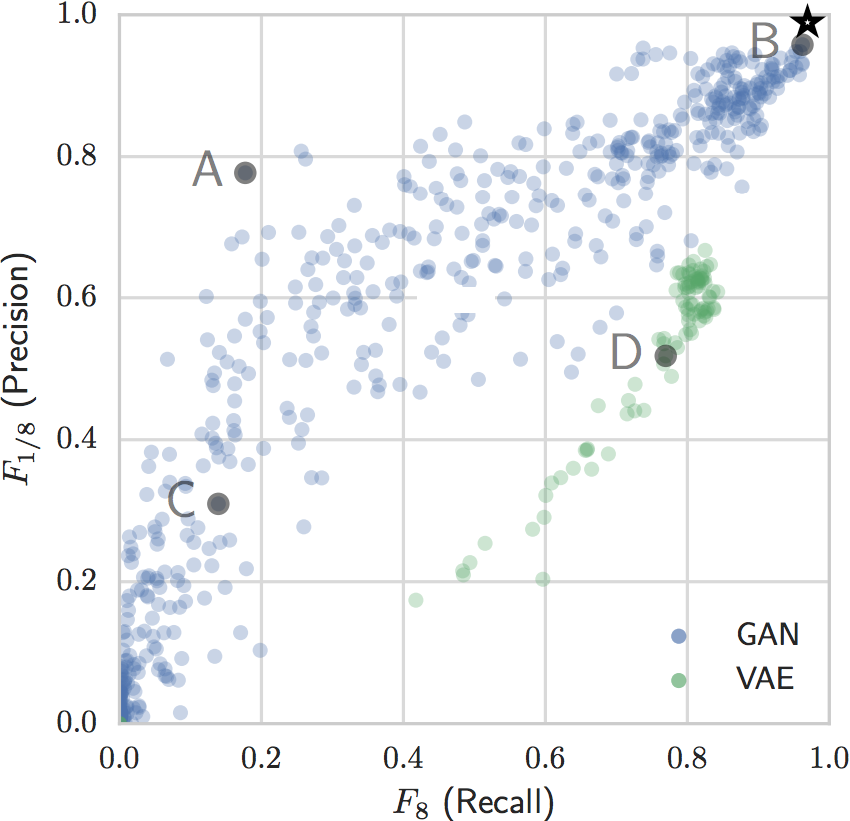} &
\includegraphics[width=0.24\linewidth]{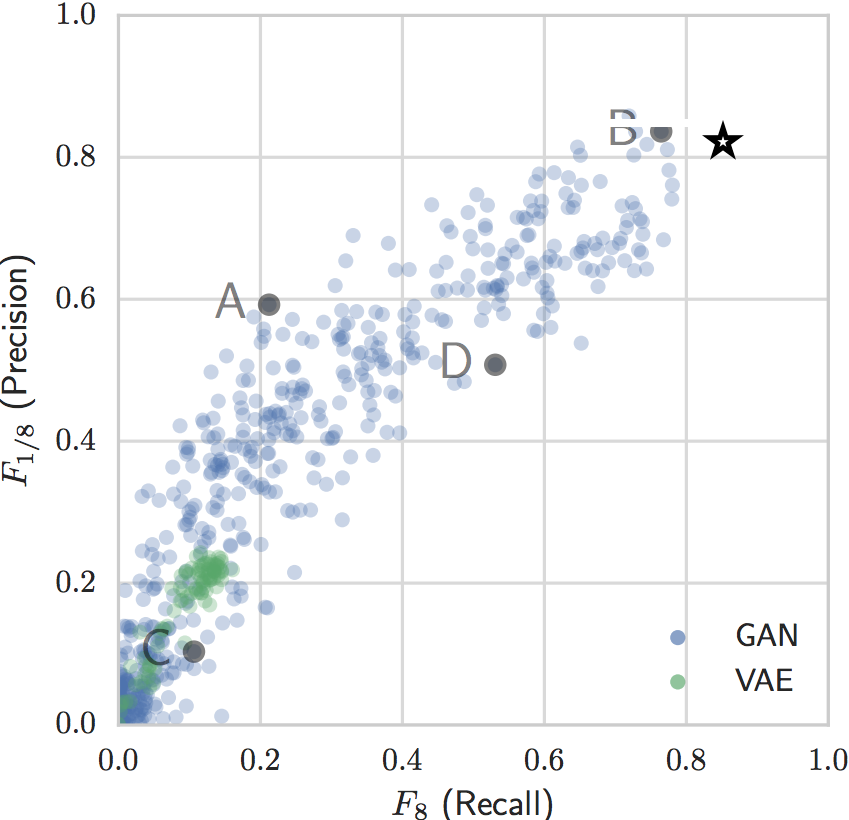} &
\includegraphics[width=0.24\linewidth]{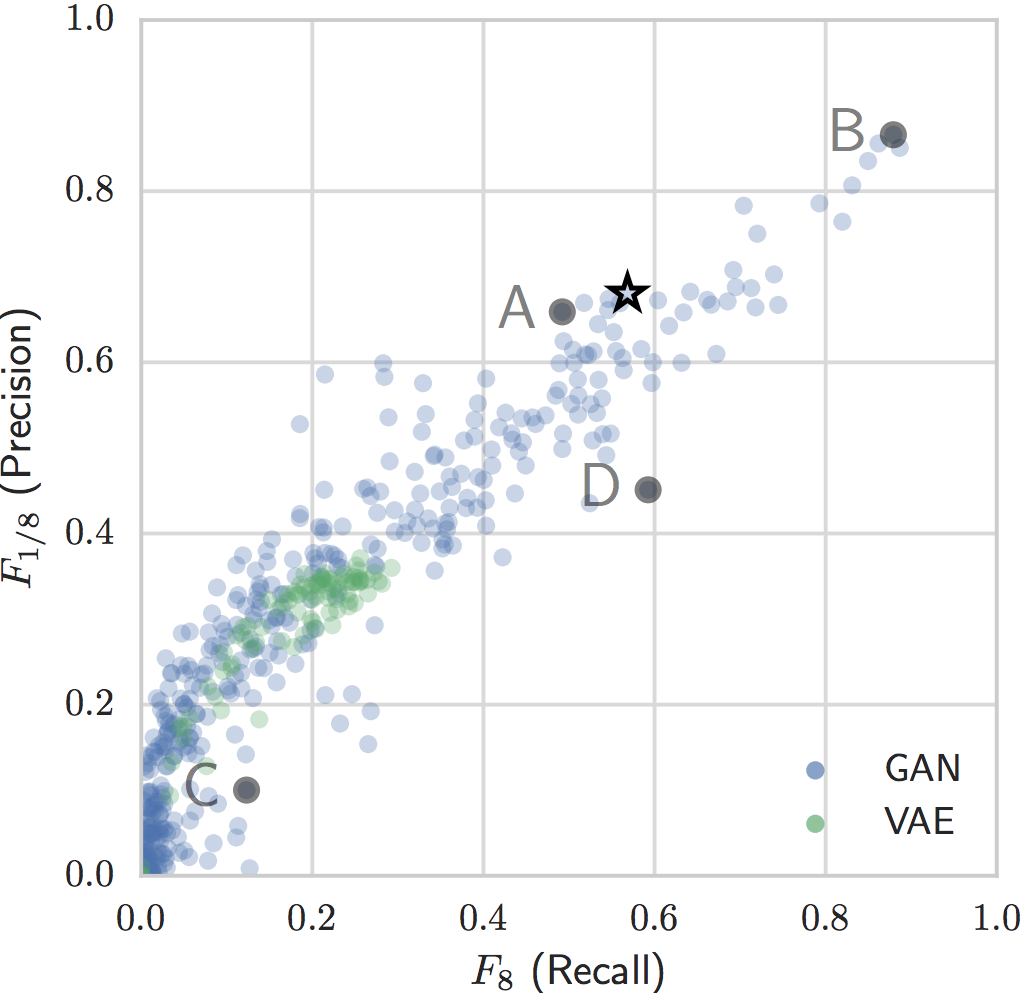} \\
    \end{tabular}

  \caption{Precision-Recall measured by $(F_8, F_{\frac{1}{8}})$ for 4 datasets. The plots were reported by \cite{sajjadi2018assessing}. We marked the results of our model for each dataset by a star on the relevant plot.}
  \label{fig:prd}
\end{figure*}

\subsection{Evaluation of Precision and Recall}
\label{sec:eval:prd}

FID is effective at measuring precision, but not recall. We therefore also opt for the evaluation metric recently presented by Sajjadi et al. \cite{sajjadi2018assessing} which they name PRD. PRD first embeds an equal number of generated and real images using the inception network. All image embeddings (real and generated) are concatenated and clustered into $B$ bins ($B=20$). Histograms $P(\omega)$, $Q(\omega)$ are computed for the number of images in each cluster from the real, generated data respectively. The precision ($\alpha$) and recall ($\beta$) are defined:

\begin{equation}
\label{eq:method:prd_a} 
\alpha(\lambda) = \sum_{\omega \in \Omega} min(\lambda P(\omega), Q(\omega))
\end{equation}

\begin{equation}
\label{eq:method:prd_b} 
\beta(\lambda) = \sum_{\omega \in \Omega} min(P(\omega), \frac{Q(\omega)}{\lambda})
\end{equation}

The set of pairs $PRD = \{(\alpha(\lambda_i), \beta(\lambda_i))\}$ forms the precision-recall curve (threshold $\lambda$ is sampled from an equiangular grid). The precision-recall curve is summarized by a variation of the $F_1$ score: $F_{\beta}$ which is able to assign greater importance to precision or recall. Specifically $(F_8, F_{\frac{1}{8}})$ are used for capturing (recall, precision).

The exact numerical precision-recall values are not available in \cite{sajjadi2018assessing}, they do provide scatter plots with the $(F_8, F_{\frac{1}{8}})$ pairs of all $800$ models trained in \cite{lucic2017gans}. We computed $(F_8, F_{\frac{1}{8}})$ for the models trained using our method as described in the previous section. The scores were computed using the authors' code. For ease of comparison, we overlay our scores over the scatter plots provided in \cite{sajjadi2018assessing}. Our numerical $(F_8, F_{\frac{1}{8}})$ scores are: MNIST $(0.971, 0.979)$, Fashion $(0.985, 0.963)$, CIFAR10 $(0.860, 0.825)$ and CelebA $(0.574, 0.681)$. The results for GLO with sampling by fitting a Gaussian to the learned latent codes (as suggested in \cite{glo}) were much worse: MNIST  $(0.845, 0.616)$, Fashion  $(0.888, 0.594)$, CIFAR10  $(0.693, 0.680)$, CelebA $(0.509, 0.404)$.

From Fig.~\ref{fig:prd} we can observe that our method generally performs better or competitively to GANs on both precision and recall. On MNIST our method and the best GAN method achieved near-perfect precision-recall. On Fashion our method achieved near perfect precision-recall while the best GAN method lagged behind. On CIFAR10 the performance of our method was also convincingly better than the best GAN model. On CelebA, our method performed well but did not achieve the top performance due to the checkerboard issue described in Sec.~\ref{sec:eval:prd}. Overall the performance of our method is typically better or equal to the baselines examined, this is even more impressive in view of the baselines being exhaustively tested over 100 hyperparameter configurations. We also note that our method outperformed VAEs and GLOs very convincingly. This provides evidence that our method is far superior to other generator-based non-adversarial models.

\begin{figure*}[t]
  \centering
    \begin{tabular}{cccc}
IMLE & GLO & GAN & Ours \\
\includegraphics[width=0.20\linewidth]{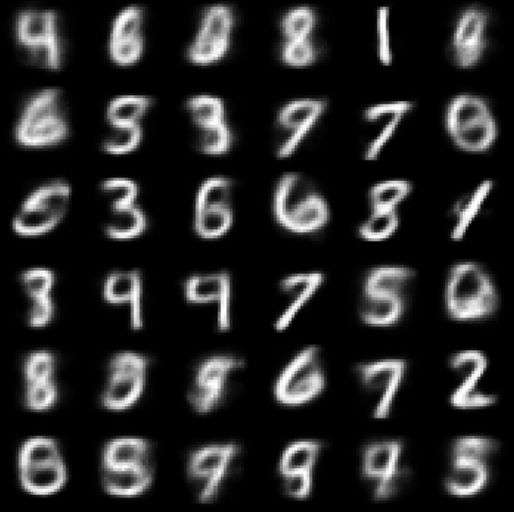} &
\includegraphics[width=0.20\linewidth]{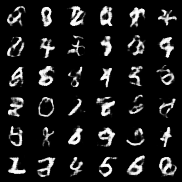} &
\includegraphics[width=0.20\linewidth]{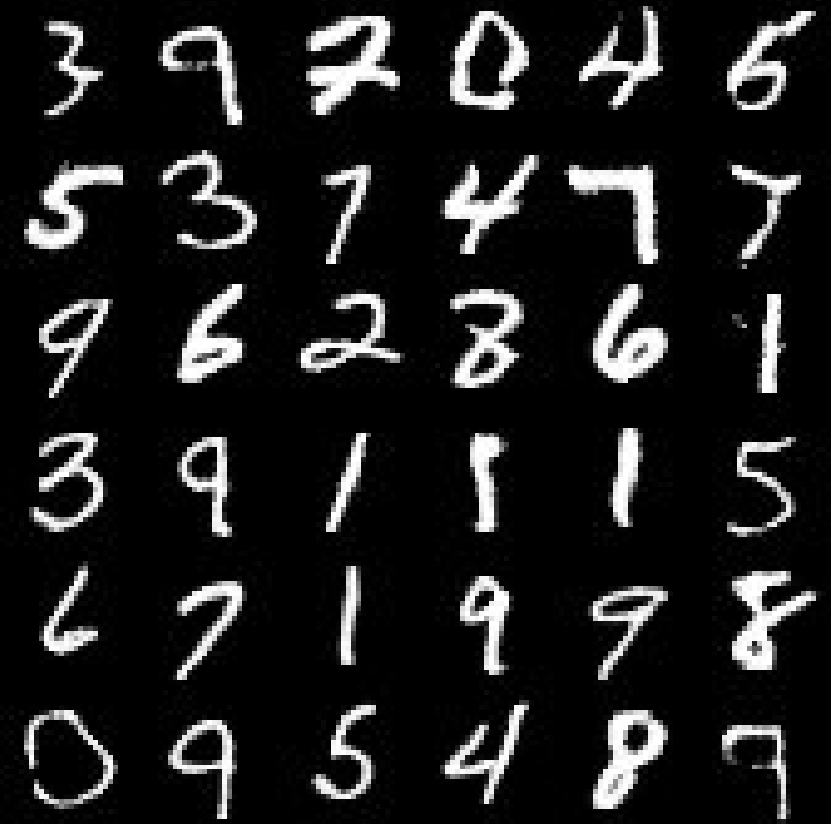} &
\includegraphics[width=0.20\linewidth]{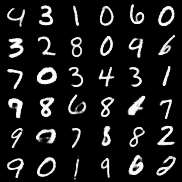} \\
&
\includegraphics[width=0.20\linewidth]{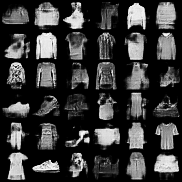}  &
\includegraphics[width=0.20\linewidth]{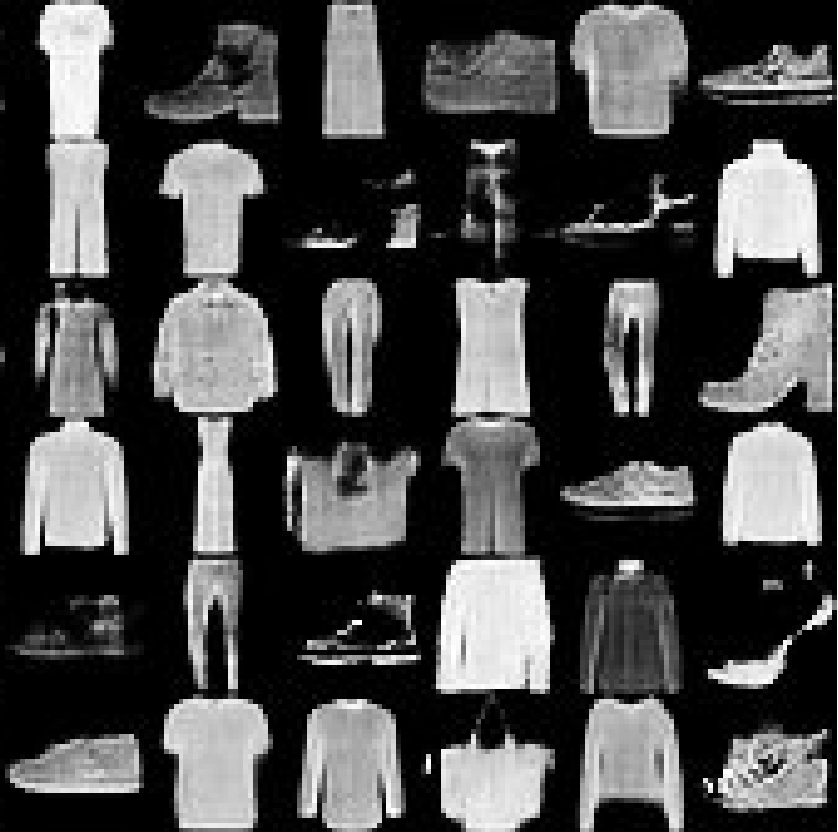} &
\includegraphics[width=0.20\linewidth]{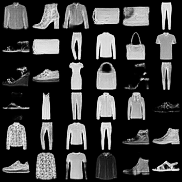} \\
\includegraphics[width=0.20\linewidth]{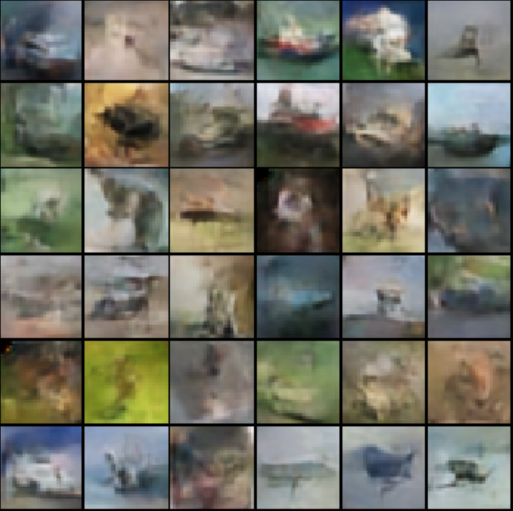} &
\includegraphics[width=0.20\linewidth]{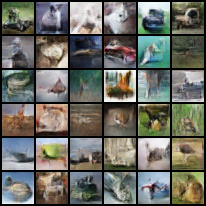} &
\includegraphics[width=0.20\linewidth]{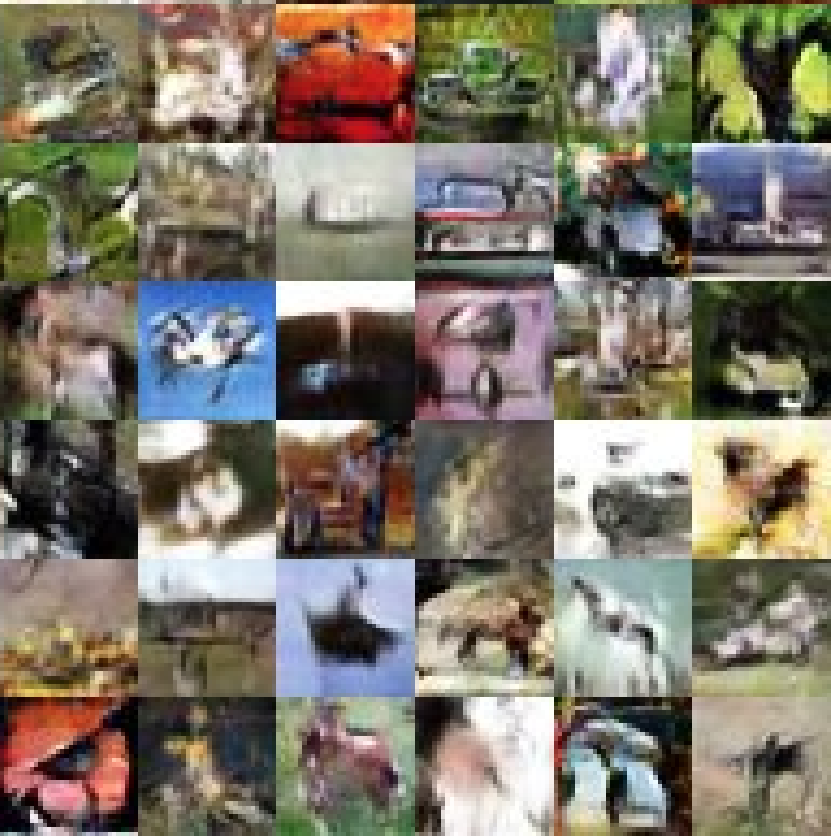} &
\includegraphics[width=0.20\linewidth]{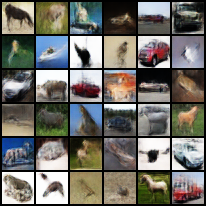} \\
&
\includegraphics[width=0.20\linewidth]{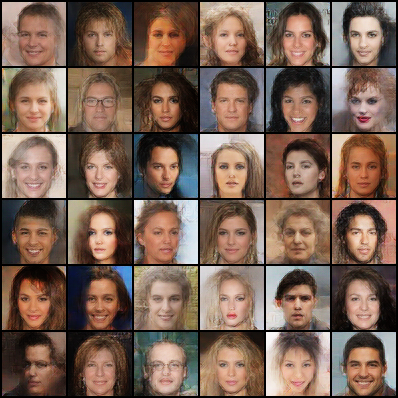} &
\includegraphics[width=0.20\linewidth]{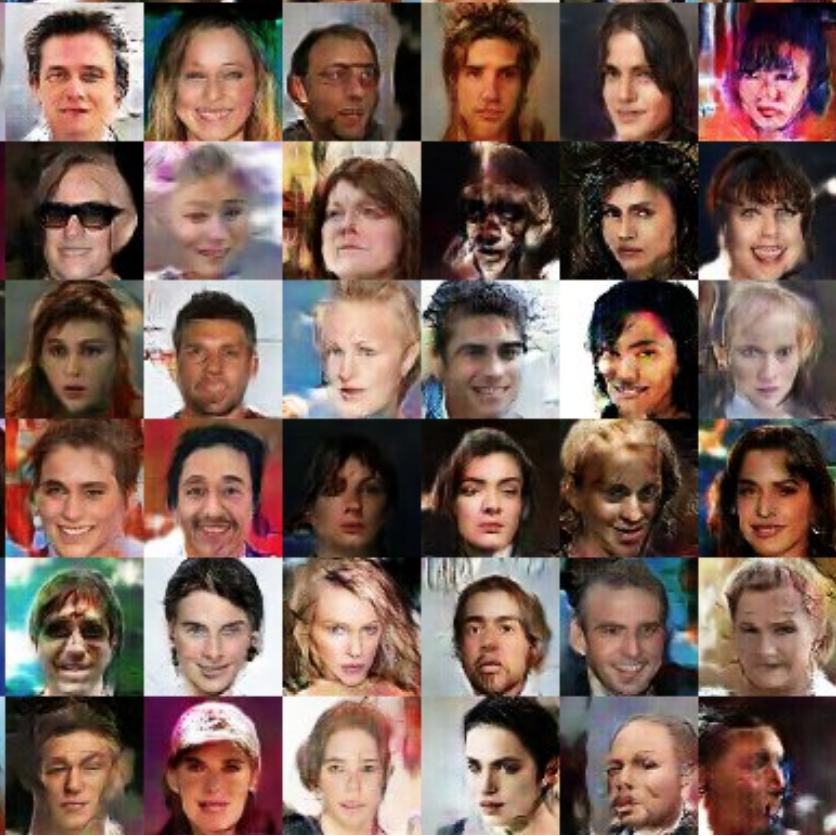} &
\includegraphics[width=0.20\linewidth]{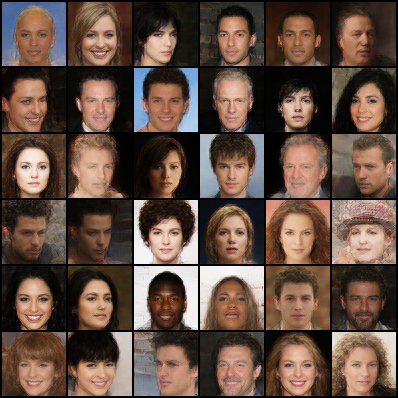} \\
    \end{tabular}

  \caption{Comparison of synthesis by IMLE \cite{imle}, GLO \cite{glo}, GAN \cite{lucic2017gans}, Ours. First row: MNIST, Second row: Fashion, Third row: CIFAR10, Last row: CelebA64. The missing IMLE images were not reported in \cite{imle}. The GAN results are taken from \cite{lucic2017gans}, corresponding to the best generative model out of $800$ as evaluated by the precision-recall metric.  }
  \label{fig:fashion}
\end{figure*}

\begin{figure*}[t]
  \centering

\includegraphics[width=0.8\linewidth]{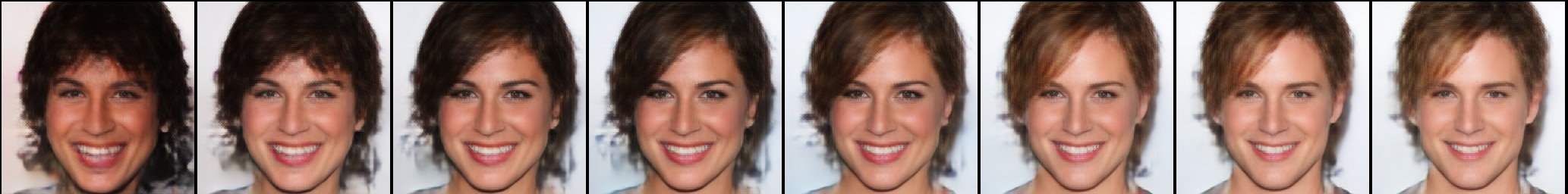} \\
\includegraphics[width=0.8\linewidth]{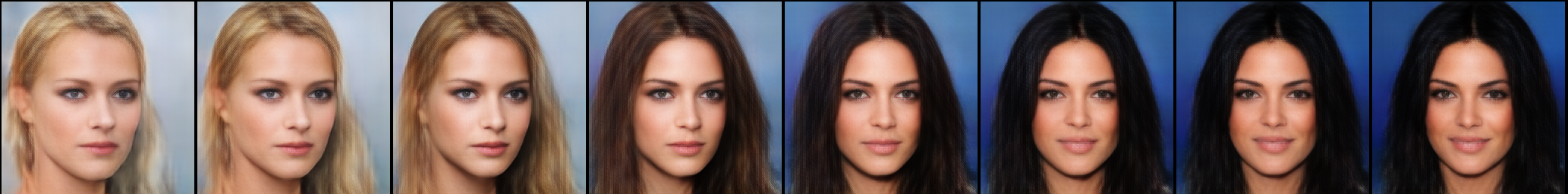} \\
\includegraphics[width=0.8\linewidth]{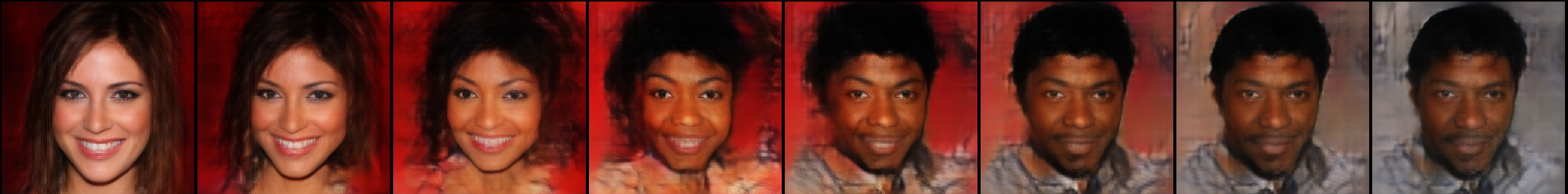} \\

  \caption{Interpolation on CelebA-HQ at $256 \times 256$ resolution. The rightmost and leftmost images are randomly sampled from random noise. The interpolation are smooth and of high visual quality.}
  \label{fig:interpolate}
\end{figure*}

\begin{figure*}[t]
  \centering

\includegraphics[width=0.8\linewidth]{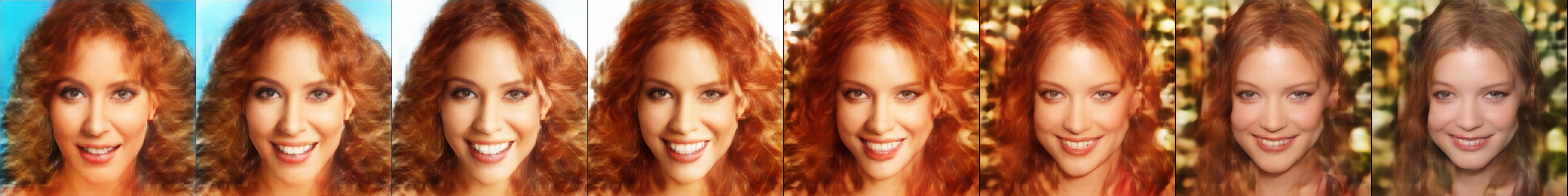} \\
\includegraphics[width=0.8\linewidth]{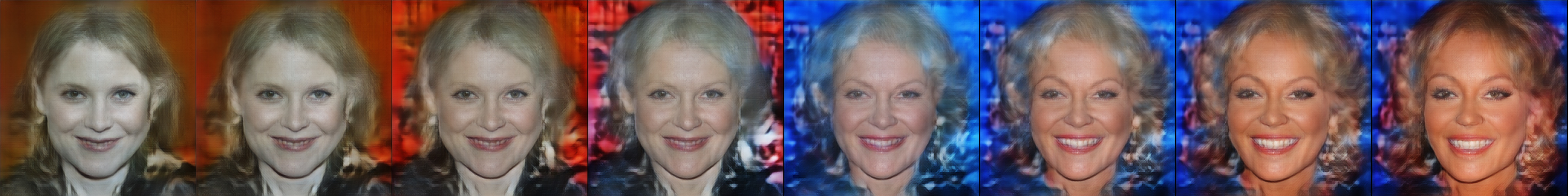} \\

  \caption{Interpolation on CelebA-HQ at $1024 \times 1024$ resolution.}
  \label{fig:interpolate1024}
\end{figure*}

\subsection{Qualitative Image Generation Results}
\label{sec:eval:qual}

We provide qualitative comparisons between our method and the GAN models evaluated by Sajjadi et al. \cite{sajjadi2018assessing}. We also show promising results on high-resolution image generation.

As mentioned above, Sajjadi et al. \cite{sajjadi2018assessing} evaluated $800$ different generative models in terms of precision and recall. They provided visual examples of their best performing model (marked as B) for each of the $4$ datasets evaluated. In Fig.~\ref{fig:fashion}, we provide a visual comparison between random samples generated by our model (without cherry picking) vs. their reported results.

We can observe that on MNIST and Fashion-MNIST our method and the best GAN method performed very well. The visual examples are diverse and of high visual quality.

On the CIFAR10 dataset, we can observe that our examples are more realistic than those generated by the best GAN model trained by \cite{lucic2017gans}. On CelebA our generated image are very realistic and with many fewer failed generations. Our generated images do suffer from some pixelization (discussed in Sec.~\ref{sec:eval:fid}). We note that GANs can generate very high quality faces (e.g. PGGAN \cite{karras2017progressive}), however it appears that for the small architecture used by Lucic et al. and Sajjadi et al., GANs do not generate particularly high-quality facial images.

To evaluate the performance of our method on higher resolution images, we trained our method on the CelebA-HQ dataset at $256 \times 256$ resolution. We used the network architecture from Mescheder et al \cite{Mescheder2018ICML}. We use $64$ channels, latent code dimensionality of $256$ and noise dimension of $100$. We used a learning rate of $0.003$ for the latent codes, $0.001$ for the generator and $0.003$ for the noise to latent code mapping function. We trained for 250 epochs, decayed by $0.5$ every 10 epochs.

We show some examples of interpolation between two randomly sampled noises in Fig.~\ref{fig:interpolate}. Several observations can be made from the figures: i) Our model is able to generate very high quality images at high resolutions. ii) The smooth interpolations illustrate that our model generalizes well to unseen images.

To show the ability of our method to scale to $1024 \times 1024$, we present  two interpolations at this high resolution in Fig.~\ref{fig:interpolate1024}, although we note that not all interpolations at such high resolution were successful.

\subsection{ModelNet Chair 3D Generation}
\label{sec:eval:3d}

\begin{figure}[t]
  \centering

\includegraphics[width=0.8\linewidth]{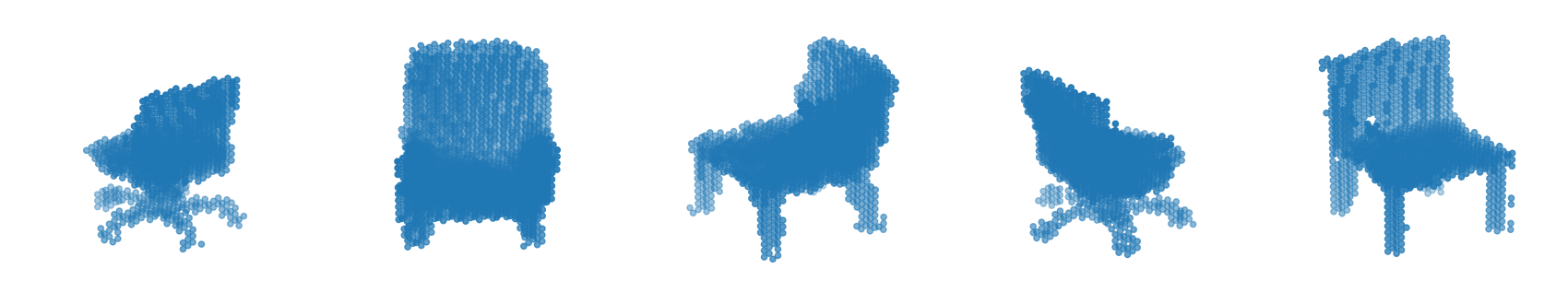}

  \caption{Examples of 3D chairs generated by GLANN}
  \label{fig:3d}
\end{figure}

To further illustrate the scope of GLANN, we present preliminary results for 3D generation on the Chairs category of ModelNet \cite{wu20153d}. The generator follows the 3DGAN architecture from \cite{3dgan}. GLANN was trained with ADAM and an $L_1$ loss. Some GLANN generated 3D samples are presented in Fig.~\ref{fig:3d}.

\subsection{Non-Adversarial Unsupervised Image Translation}
\label{sec:eval:mapping}

As generative models are trained in order to be used in downstream tasks, we propose to evaluate generative models by the downstream task of cross domain unsupervised mapping. NAM \cite{nam_eccv} was proposed by Hoshen and Wolf for unsupervised domain mapping. The method relies on having a strong unconditional generative model of the output image domain.  Stronger generative models perform better at this task. This required \cite{nam_eccv,vaenam} to use GAN-based unconditional generators. We evaluated our model using the $3$ quantitative benchmarks presented in \cite{nam_eccv} - namely: $MNIST \rightarrow SVHN$, $SVHN \rightarrow MNIST$ and $Car \rightarrow Car$. Our model achieved scores of $31.3\%$, $25.0\%$ and $1.45$ on the three tasks respectively. The results are similar to those obtained using the GAN-based unconditional models (although SVHN is a bit lower here). GLANN is therefore the first model able to achieve fully unsupervised image translation without the use of GANs.

\section{Discussion}
\label{sec:disc}

\textbf{Loss function:} In this work, we replaced the standard adversarial loss function by a perceptual loss. In practice we use ImageNet-trained VGG features. Zhang et al. \cite{zhang2018unreasonable} claimed that self-supervised perceptual losses work no worse than the ImageNet-trained features. It is therefore likely that our method will have similar performance with self-supervised perceptual losses.

\textbf{Higher resolution:} The increase in resolution between $64 \times 64$ to $256 \times 256$ or $1024 \times 1024$ was enabled by a simple modification of the loss function: the perceptual loss was calculated both on the original images, as well as on a bi-linearly subsampled version of the image. Going up to higher resolutions simply requires more sub-sampling levels. Research into more sophisticated perceptual loss will probably yield further improvements in synthesis quality.

\textbf{Other modalities:} In this work we focuses on image synthesis. We believe that our method can extend to many other modalities, particularly 3D and video. The simplicity of the procedure and robustness to hyperparameters makes application to other modalities much simpler than GANs. We showed some evidence for this assertion in Sec.~\ref{sec:eval:3d}. One research task for future work is finding good perceptual loss functions for domains outside 2D images.  

\section{Conclusions}
\label{sec:conc}

In this paper we introduced a novel non-adversarial method for training generative models. Our method combines ideas from GLO and IMLE and overcomes the weaknesses of both methods. When compared on established benchmarks, our method outperformed the the most common GAN models that underwent exhaustive hyperparameter tuning. Our method is robust and simple to train and achieves excellent results. As future work, we plan to extend this work to higher resolutions and new modalities such as video and 3D.

{\small
\bibliographystyle{ieee}
\bibliography{egbib}
}

\end{document}